\documentclass[11pt]{article}

\usepackage[final]{acl}

\usepackage{times}
\usepackage{latexsym}

\usepackage[T1]{fontenc}

\usepackage[utf8]{inputenc}

\usepackage{microtype}

\usepackage{inconsolata}

\usepackage{graphicx}
\usepackage{algorithm}%
\usepackage{algorithmicx}%
\usepackage{algpseudocode}%
\usepackage{listings}%
\usepackage{booktabs}
\usepackage{amsmath}
\usepackage{amssymb}
\usepackage{amsfonts}
\usepackage{multirow}
\usepackage{subfig}
\usepackage[table]{xcolor}

\title{Balancing Global Quality and Pronoun-Specific Feedback for Context-Aware Machine Translation}

\author{
  \textbf{Harshit Dhankhar\textsuperscript{1}\thanks{Work done while at Indian Institute of Technology Patna.}},
  \textbf{Baban Gain\textsuperscript{2}},
  \textbf{Asif Ekbal\textsuperscript{2}},  \textbf{Yogesh Mani Tripathi\textsuperscript{3}}
  \\
  \textsuperscript{1}Department of Computer Science \& Engineering, University of California San Diego, USA
  \\
  \textsuperscript{2}Department of Computer Science \& Engineering,
  Indian Institute of Technology Patna, India
  \\
  \textsuperscript{3}Department of Mathematics,
  Indian Institute of Technology Patna, India
}
\begin{document}
\maketitle

\begin{abstract}
Context-aware machine translation can expose the evidence needed for pronoun choice, but standard fine-tuning does not explicitly prioritize these sparse discourse-sensitive decisions. We study \textsc{ProNMT}, a reward-guided iterative self-training method that combines sentence-level quality estimation with a signed confidence signal at generated pronoun positions. For each current sentence and its preceding source context, \textsc{ProNMT} samples candidate translations, scores them using reference-free quality estimation together with a reference-derived pronoun label, and fine-tunes on the highest-scoring candidate. On filtered English--German Europarl and English--French News Commentary data, \textsc{ProNMT} improves over context-aware supervised fine-tuning on BLEU and COMET. Ablations show that pronoun-only feedback can severely degrade sentence-level translation quality on these pronoun-focused data, while hard binary feedback underperforms confidence-weighted feedback. These results indicate that targeted linguistic feedback is most useful when combined with both a global quality signal and the context relevant to the targeted decision. We make the code publicly available at \url{https://github.com/Harshit2807161/ProNMT}.
\end{abstract}

\section{Introduction}
\begin{figure}[t]
    \centering
    \includegraphics[width=0.48\textwidth]{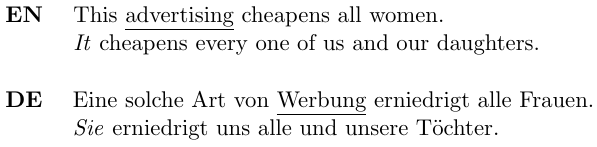}
    \caption{A preceding source sentence provides evidence for pronoun choice in the current sentence. Pronouns are italicized and antecedents are underlined. Data: Europarl English--German.}
    \label{fig:example1}
\end{figure}

Pronoun translation is a compact but difficult test of context-aware machine translation. A pronoun in the current sentence may be ambiguous without an antecedent or discourse cue from a preceding sentence, and a system that accepts additional context does not necessarily use that context for the relevant decision \cite{li-etal-2020-multi-encoder,appicharla-etal-2023-case}. Figure~\ref{fig:example1} illustrates an English--German example in which the translation of \textit{it} depends on the antecedent \textit{advertising} in the preceding sentence.

Context-aware architectures and training objectives improve access to information beyond the sentence boundary \cite{tiedemann-scherrer-2017-neural,junczys-dowmunt-2019-microsoft,voita-etal-2018-context}. However, broad sentence-level objectives do not explicitly prioritize sparse linguistic decisions such as pronoun choice. Quality estimation (QE) offers a practical global signal for feedback-guided MT training because it can score candidate translations without human judgments or a reference at scoring time \cite{rei-etal-2021-references,he-etal-2024-improving}. Yet a global QE score may overlook a locally important pronoun error.

A natural response is to add a pronoun-specific signal. Such a signal introduces a second problem: optimizing a narrow local objective may reward candidates that satisfy the pronoun criterion while damaging the rest of the translation. The interaction among context, global quality, and targeted feedback is therefore central. We study this interaction through \textsc{ProNMT}, a reward-guided iterative self-training framework. The translation model samples candidates, a combined reward ranks them, and the highest-scoring candidate becomes the target for a supervised update. The reward contains two components: reference-free QE for global translation quality and a reference-conditioned signed confidence signal at a generated pronoun position.

Our experiments attempt to answer the following four questions:
\begin{enumerate}
    \item Does pronoun-specific feedback add value beyond context-aware supervised fine-tuning?
    \item Are global QE and local pronoun feedback complementary?
    \item Is the local feedback more useful when preceding-sentence context is available?
    \item Does confidence weighting rank candidates more effectively than a hard binary pronoun reward?
\end{enumerate}

Our contributions are threefold. First, we formulate pronoun-aware QE-guided self-training that combines a global reference-free quality signal with reference-derived pronoun supervision. Second, we provide a controlled comparison of context-aware and context-agnostic conditions together with QE-only, pronoun-only, combined, and binary-reward variants. Third, experiments on English--German and English--French show that local pronoun feedback is harmful in isolation but complementary to global QE when the model can access potentially relevant preceding context.

\section{Related Work}
\subsection{Context-Aware and Pronoun-Focused MT}
Context-aware MT commonly incorporates preceding sentences through concatenation \cite{tiedemann-scherrer-2017-neural,junczys-dowmunt-2019-microsoft}, multiple encoders \cite{li-etal-2020-multi-encoder}, or auxiliary objectives \cite{appicharla-etal-2024-case}. Prior work shows that context can improve anaphora-sensitive translation \cite{voita-etal-2018-context}, but also that models may remain insensitive to whether the supplied context is relevant \cite{li-etal-2020-multi-encoder,appicharla-etal-2023-case}. Preceding-sentence information has also been used in multilingual and document-aware MT systems \cite{wang-etal-2020-tencent,sun-etal-2022-rethinking,wang-etal-2023-document-level}.

ContraPro provides a large contrastive test suite for English--German pronoun translation \cite{mueller2018}. It is designed primarily for controlled evaluation rather than as a natural parallel corpus for iterative candidate-selection training. We therefore train on filtered parallel corpora and analyze the preliminary ContraPro training attempt in Section~\ref{sec:contra}.

\subsection{Feedback-Guided Training for MT}
MT systems have been improved using human bandit feedback and other limited-feedback settings \cite{kreutzer-etal-2018-neural,kreutzer-etal-2018-reliability}. Reinforcement learning from human feedback can optimize learned preferences, but commonly requires human annotations and additional optimization components \cite{10.5555/3600270.3602281,santacroce2023efficient,ramamurthy2023reinforcementlearningnotnatural}. In contrast, our procedure does not optimize a policy with PPO. It uses reward-based candidate selection followed by iterative supervised updates.

\subsection{Quality Estimation as a Training Signal}
QE estimates translation quality without a reference at inference time \cite{rei-etal-2020-unbabels,rei-etal-2021-references}. Recent work has used QE as a reward model for MT improvement \cite{he-etal-2024-improving}. Our focus is complementary: we test whether a global QE signal benefits from an additional linguistically targeted signal for a sparse contextual decision, and whether that local signal remains safe when optimized jointly with sentence-level quality.

\section{Pronoun-Aware QE-Guided Self-Training}
\subsection{Task and Context Representation}
Let $x_i$ be the current source sentence and $x_{i-1}$ its preceding source sentence. In the context-aware condition, the translation model receives
\begin{equation}
\tilde{x}_i = \langle\mathrm{context}\rangle\, x_{i-1}\, \langle/\mathrm{context}\rangle\, x_i,
\end{equation}
while the desired output is the translation of $x_i$ only. In the context-agnostic condition, the model receives $x_i$. The target reference $y_i^{\mathrm{ref}}$ supplies the pronoun label used by the local reward. Thus, only the QE component is reference-free; the complete training objective combines reference-free QE with reference-derived pronoun supervision.

\subsection{Candidate Selection and Iterative Updates}
Let $M_i=P(y\mid x;\theta_i)$ denote the translation model at iteration $i$. For every input in a mini-batch, we sample up to $k$ distinct candidate translations using multinomial sampling. Each candidate is scored by the reward in Equation~\ref{eq:combined_reward}. The highest-scoring candidate is then used as the target for an iterative supervised fine-tuning update. Algorithm~\ref{alg:pronmt} summarizes the procedure.

\begin{algorithm}[t]
\caption{\textsc{ProNMT} iterative self-training}
\label{alg:pronmt}
\begin{algorithmic}[1]
\Require Training inputs $\mathcal{X}$, references $\mathcal{Y}$, initial model $M_0$, reward $r$, batch size $B$, temperature $T$, candidate limit $k$
\For{$i=0,1,\ldots,N-1$}
    \State $D_i \gets \mathrm{SampleBatch}(\mathcal{X},\mathcal{Y},B)$
    \State $\mathcal{B} \gets \emptyset$
    \For{each $(x,y^{\mathrm{ref}})\in D_i$}
        \State $y_1,\ldots,y_k \sim P_T(y\mid x;\theta_i)$
        \State $y^* \gets \arg\max_{y_j} r(x,y_j,y^{\mathrm{ref}})$
        \State $\mathcal{B} \gets \mathcal{B}\cup\{(x,y^*)\}$
    \EndFor
    \State Update $\theta_i$ on $\mathcal{B}$ to obtain $M_{i+1}$
\EndFor
\end{algorithmic}
\end{algorithm}

This is reward-guided self-training rather than reinforcement learning: the reward ranks sampled candidates, but the model update is a supervised update on the selected candidate.

\subsection{Global Translation-Quality Signal}
For a candidate $y$, $R_{\mathrm{QE}}(x_i,y)$ is produced by a reference-free COMET-QE model. In the context-aware condition, the translation model receives $\tilde{x}_i$, but QE scores the candidate against the current source sentence $x_i$. This prevents the context prefix from being treated as content that should appear in the translation.

\subsection{Signed Pronoun-Position Confidence}
For a generated candidate $y=(y_1,\ldots,y_m)$, let $j$ be a position at which the decoded candidate contains a recognized target-language pronoun. We define pronoun-position confidence as
\begin{equation}
C_{\mathrm{pro}}(y,j)=\max_{v\in\mathcal{V}}P(v\mid x,y_{<j};\theta),
\end{equation}
where $\mathcal{V}$ is the target vocabulary. This quantity measures model confidence at the position where a pronoun is generated; it is not an independent pronoun-accuracy metric.

Let $p(y)$ be the generated pronoun and $p^{\mathrm{ref}}$ the aligned reference pronoun. If such a pronoun is not generated we reward that case 0. The signed local reward is then:
\begin{equation}
R_{\mathrm{pro}}(y)=
\begin{cases}
+C_{\mathrm{pro}}(y,j), & p(y)=p^{\mathrm{ref}},\\
-C_{\mathrm{pro}}(y,j), & p(y)\neq p^{\mathrm{ref}},\\
0, & \text{otherwise}
\end{cases}
\label{eq:pronoun_reward}
\end{equation}
The combined candidate-selection reward is
\begin{equation}
\begin{split}
r(x_i,y,y_i^{\mathrm{ref}})={}&\alpha R_{\mathrm{QE}}(x_i,y)\\
&+\beta R_{\mathrm{pro}}(y).
\end{split}
\label{eq:combined_reward}
\end{equation}
QE provides a global constraint on sentence-level translation quality, while $R_{\mathrm{pro}}$, also referred to as ProConf when averaged across multiple sentences, prioritizes the targeted pronoun decision.

\subsection{Binary Pronoun Reward}
To test whether confidence weighting is useful, we replace Equation~\ref{eq:pronoun_reward} with
\begin{equation}
R_{\mathrm{bin}}(y)=
\begin{cases}
+1, & p(y)=p^{\mathrm{ref}},\\
-1, & p(y)\neq p^{\mathrm{ref}},\\
0, & \text{otherwise.}
\end{cases}
\end{equation}
The binary variant assigns the same magnitude to every matching or non-matching candidate. The confidence-weighted variant therefore provides a finer ranking among candidate translations.
\begin{table}
\centering
\caption{Source-sentence length statistics for ContraPro and filtered Europarl.}
\label{tab:length_stats}
\resizebox{\columnwidth}{!}{%
\begin{tabular}{lrrrrrr}
\toprule
\textbf{Dataset} & \textbf{Mean} & \textbf{Median} & \textbf{Var.} & \textbf{Std.} & \textbf{Max} & \textbf{Min}\\
\midrule
ContraPro & 12.533 & 11.0 & 53.031 & 7.282 & 67 & 2\\
Europarl & 33.378 & 29.0 & 404.219 & 20.105 & 212 & 1\\
\bottomrule
\end{tabular}%
}
\end{table}
\section{Experimental Setup}
\subsection{Pronoun-Focused Parallel Corpora}
For English--German, we start from 1,920,209 Europarl sentence pairs. We retain a pair when the English sentence contains \textit{it} exactly once, the German sentence contains exactly one singular third-person pronoun from \textit{er}, \textit{sie}, and \textit{es}, and the two pronouns are aligned by \texttt{fast\_align} \cite{dyer-etal-2013-simple}. Alignment filtering reduces noise from alternatives such as \textit{das} and \textit{dies} \cite{mueller2018}. The filtered data contains 117,834 \textit{es}, 17,447 \textit{er}, and 39,439 \textit{sie} instances. We sample 15,000 instances per class and construct 42,750 training, 750 validation, and 1,500 test examples.

For English--French, we apply the same procedure to News Commentary v16 from OPUS \cite{tiedemannopus}, retaining English \textit{they} and its French counterparts \textit{ils} and \textit{elles}. The English--German setting therefore targets grammatical-gender alternatives for \textit{it}, while English--French targets the masculine and feminine alternatives of \textit{they}. The English--German reward weights are transferred to English--French without retuning.

For each corpus, the context-aware input prefixes the current source sentence with the preceding source sentence. The same data split and random seed are used for corresponding context-aware and context-agnostic conditions.

\subsection{Choice of Training Data and ContraPro Analysis}
\label{sec:contra}
\begin{figure}[t]
    \centering
    \subfloat[ContraPro]{\includegraphics[width=0.9\columnwidth]{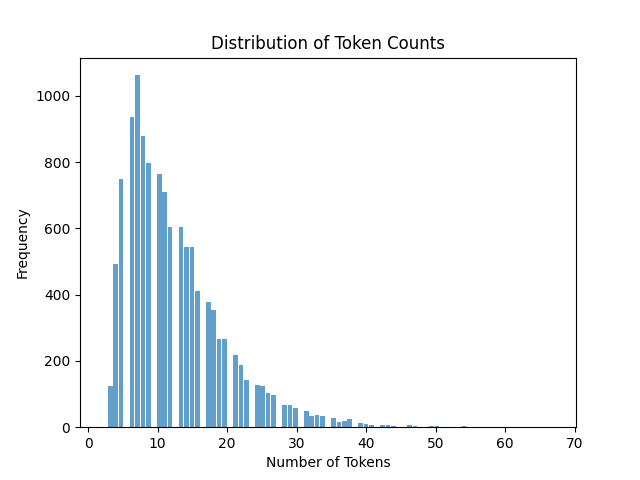}}
    \\
    \subfloat[Filtered Europarl]{\includegraphics[width=0.9\columnwidth]{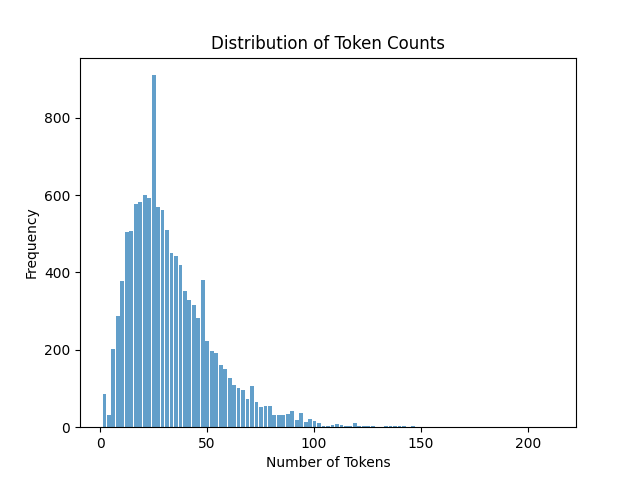}}
    \caption{Source token-count distributions.}
    \label{fig:data_comp}
\end{figure}

\begin{figure}[t]
    \centering
    \subfloat[QE reward]{\includegraphics[width=0.9\columnwidth]{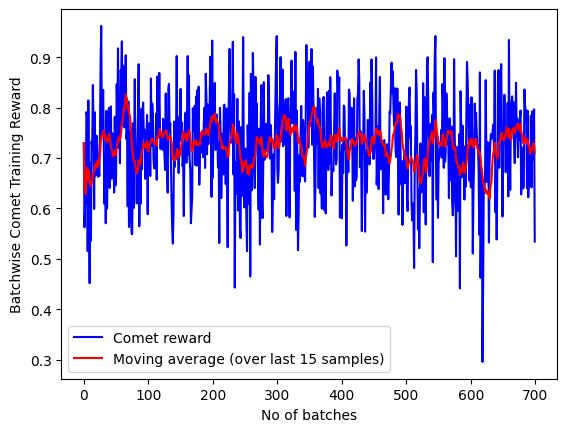}}
    \\
    \subfloat[Pronoun reward]{\includegraphics[width=0.9\columnwidth]{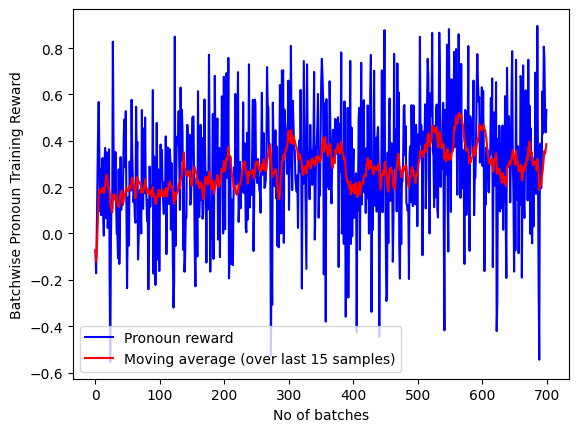}}
    \caption{Preliminary ContraPro training curves over the first 700 updates.}
    \label{fig:contra_curves}
\end{figure}

ContraPro is a controlled contrastive evaluation suite rather than a natural parallel corpus designed for iterative self-training. We nevertheless conducted a preliminary training attempt. The source sentences are shorter on average than those in our filtered Europarl corpus, as summarized in Table~\ref{tab:length_stats} and Figure~\ref{fig:data_comp}. Before training, the average validation COMET-QE score was 0.7636 on ContraPro and 0.6323 on filtered Europarl. This left less QE headroom in the ContraPro training attempt (see Figure \ref{fig:contra_curves}). Sentence length may contribute to this difference, but the current experiments do not establish a causal explanation.

\subsection{Model and Training}
All experiments use the distilled 600M-parameter NLLB-200 model. Before reward-guided training, the context-aware model is adapted through supervised fine-tuning so that it learns to consume the context prefix while translating only the current sentence. This context-aware SFT model is the principal baseline for evaluating the subsequent reward-guided training stage after adaptation to the context-prefixed input format. We do not include a separately continued SFT control for the context-agnostic setting. Consequently, comparisons between context-aware and context-agnostic \textsc{ProNMT} should be interpreted as comparisons between the two training conditions rather than as a controlled estimate of the effect of preceding context alone.

For \textsc{ProNMT}, the mini-batch size is $B=4$, and up to $k=10$ distinct candidates are generated for each input by multinomial sampling. Iterative supervised training runs for at most 700 updates with learning rate $6\times10^{-5}$ and gradient accumulation of 16. The binary-reward comparison uses the same setup and is trained for 750 updates. The supervised adaptation and reward-guided stages use the settings summarized below.

\begin{table}[t]
\centering
\caption{Hyperparameters for initial context-aware supervised fine-tuning.}
\label{tab:context_sft_hparams}
\begin{tabular}{ll}
\toprule
\textbf{Hyperparameter} & \textbf{Value}\\
\midrule
Learning rate & $8\times10^{-3}$\\
Per-device train batch size & 8\\
Gradient accumulation steps & 16\\
Evaluation accumulation steps & 16\\
Per-device evaluation batch size & 8\\
Weight decay & 0.01\\
Number of epochs & 10\\
\bottomrule
\end{tabular}
\end{table}

\subsection{Compared Systems}
We compare the following configurations:
\begin{itemize}
    \item \textbf{Baseline}: the NLLB-200 model before task-specific reward-guided training;
    \item \textbf{Context-aware SFT}: supervised adaptation to the context-prefixed input format;
    \item \textbf{QE only}: candidate selection using $R_{\mathrm{QE}}$;
    \item \textbf{Pronoun only}: candidate selection using $R_{\mathrm{pro}}$;
    \item \textbf{\textsc{ProNMT}}: joint QE and confidence-weighted pronoun feedback;
    \item \textbf{Binary}: joint QE and binary pronoun feedback.
\end{itemize}

\subsection{Validation and Evaluation}
For each reward-weight configuration, checkpoints are evaluated every 100 updates, and the checkpoint with the highest combined validation reward is retained. The reward weights $\alpha$ and $\beta$ are then selected using COMET and BLEU on the validation set. Table~\ref{tab:weight_sensitivity} reports the corresponding English--German validation results. The held-out test set is used only for the final evaluations reported in Tables~\ref{tab:ende_main} and~\ref{tab:enfr_main}.

We report BLEU \cite{papineni-etal-2002-bleu} and reference-based COMET using \textit{wmt22-comet-da} \cite{rei-etal-2020-comet}. COMET is presented on a 0--100 scale. We also report the mean QE reward and mean signed pronoun-position confidence used for candidate selection. These internal reward values are diagnostic quantities, not independent measures of pronoun accuracy. English--German and English--French QE values are not directly comparable because the two experiments use different QE models.

\begin{table*}[t]
\centering
\caption{English--German validation performance under different reward weights. $\#\mathrm{tokens}$ is the sentence-specific token count and $\#\mathrm{avg\mbox{-}len}\approx 30$ is the average source length. QE and ProConf are validation-set internal rewards.}
\label{tab:weight_sensitivity}
\resizebox{0.65\linewidth}{!}{%
\begin{tabular}{lrrrr}
\toprule
\textbf{Reward weights $(\alpha,\beta)$} & \textbf{COMET} & \textbf{BLEU} & \textbf{QE} & \textbf{ProConf}\\
\midrule
\multicolumn{5}{c}{\textit{Without preceding context}}\\ \midrule
$(1,1/\#\mathrm{tokens})$ & 72.88 & 15.2200 & \textbf{0.1189} & \textbf{0.3769}\\
$(1,1/\#\mathrm{avg\mbox{-}len})$ & \textbf{73.27} & \textbf{15.2318} & 0.1026 & 0.3524\\
$(1.2,1/\#\mathrm{tokens})$ & 63.30 & 10.8723 & 0.0069 & 0.3225\\
$(1.2,1/\#\mathrm{avg\mbox{-}len})$ & 64.03 & 11.5233 & 0.0551 & 0.3115\\
\midrule
\multicolumn{5}{c}{\textit{With preceding context}}\\ \midrule
$(1,1/\#\mathrm{tokens})$ & 78.66 & 21.2250 & 0.0127 & 0.8270\\
$(1,1/\#\mathrm{avg\mbox{-}len})$ & 78.98 & 21.9542 & 0.0121 & \textbf{0.8543}\\
$(1.2,1/\#\mathrm{tokens})$ & \textbf{80.72} & \textbf{26.4251} & 0.0223 & 0.4213\\
$(1.2,1/\#\mathrm{avg\mbox{-}len})$ & 80.21 & 25.2356 & \textbf{0.0311} & 0.4246\\
\bottomrule
\end{tabular}%
}
\end{table*}

\section{Results and Analysis}
\subsection{Does Reward-Guided Training Improve over Context-Aware SFT?}
Table~\ref{tab:ende_main} reports English--German test results. Context-aware SFT reaches 81.19 COMET and 25.57 BLEU. Joint QE and pronoun feedback improves these scores to 81.92 COMET and 26.96 BLEU, gains of 0.73 COMET and 1.39 BLEU. Although the COMET difference is modest, the improvement is consistent across both a learned reference-based metric and an overlap-based metric. The combined system also raises the internal pronoun-position confidence from 0.1406 to 0.4183.  We do not treat this internal increase as an independent accuracy result, but it confirms that the selected candidates are more strongly aligned with the targeted reward while external translation quality is preserved.

The comparison with context-aware SFT distinguishes the initial adaptation to the context-prefixed input format from the subsequent reward-guided refinement stage. However, because \textsc{ProNMT} receives additional parameter updates after SFT, this comparison does not by itself isolate the effect of reward-guided candidate selection from that of additional training. We therefore interpret the gain over SFT as evidence that the complete \textsc{ProNMT} training procedure further improves the adapted model, rather than as a controlled estimate of the contribution of candidate selection alone.

The English--German improvement should nevertheless be interpreted as a refinement of an already strong system rather than a replacement for supervised context adaptation. Context-aware SFT accounts for most of the improvement over the raw context baseline, while the subsequent \textsc{ProNMT} stage is associated with a further improvement. This layered interpretation is more faithful to the training pipeline: SFT first establishes the ability to consume the extended input, and reward-guided self-training then encourages the model to prefer candidates that jointly satisfy sentence-level quality and the targeted pronoun criterion.

\begin{table*}[t]
\centering
\caption{English--German test performance. QE and ProConf are the mean internal QE reward and signed pronoun-position confidence. The reward weights are selected on the validation set.}
\label{tab:ende_main}
\resizebox{0.6\linewidth}{!}{%
\begin{tabular}{lrrrr}
\toprule
\textbf{Method} & \textbf{COMET} & \textbf{BLEU} & \textbf{QE} & \textbf{ProConf}\\
\midrule
\multicolumn{5}{c}{\textit{Without preceding context}}\\ \midrule
Baseline & 56.68 & 7.8904 & -0.1154 & 0.0902\\
Pronoun only & 50.30 & 1.8290 & -0.1534 & \textbf{0.8998}\\
QE only & 58.30 & 2.6723 & 0.0480 & 0.3798\\
\textsc{ProNMT} & \textbf{73.95} & \textbf{15.2630} & \textbf{0.1186} & 0.3672\\
Binary & 63.02 & 12.2501 & 0.1052 & 0.7500\\
\midrule
\multicolumn{5}{c}{\textit{With preceding context}}\\ \midrule
Baseline & 66.03 & 11.5783 & 0.0598 & 0.3370\\
Context-aware SFT & 81.19 & 25.5680 & 0.0258 & 0.1406\\
Pronoun only & 16.79 & 0.0703 & -0.6024 & \textbf{0.9675}\\
QE only & 80.06 & 21.5588 & 0.0303 & 0.2262\\
\textsc{ProNMT} & \textbf{81.92} & \textbf{26.9574} & \textbf{0.0362} & 0.4183\\
Binary & 78.58 & 21.5808 & 0.0173 & 0.7500\\
\bottomrule
\end{tabular}%
}
\end{table*}

\subsection{Why Are Both Reward Components Necessary?}
The clearest result is the failure of pronoun-only optimization. In the context-aware English--German condition, pronoun-only training reaches a high internal ProConf value of 0.9675 but collapses to 16.79 COMET and 0.07 BLEU. The same conflict appears without context: pronoun-only training raises ProConf to 0.8998, yet COMET and BLEU fall below even the unadapted baseline. The local objective can therefore be satisfied while the rest of the translation deteriorates. ProConf should not be interpreted as pronoun accuracy, especially when considered without global translation metrics.

This failure is a candidate-selection analogue of reward over-optimization. The local term examines a single targeted position, so a candidate can receive a favorable local score even when its lexical choice, word order, coverage, or fluency elsewhere is poor. Repeatedly training on such candidates can amplify the mismatch between the optimized signal and the broader translation task. The near-zero BLEU score in the context-aware pronoun-only condition shows that this is not a minor trade-off. It is a qualitative collapse of the sentence-level objective. The result motivates using the pronoun signal as a preference among otherwise plausible translations, rather than as a standalone definition of quality.

QE-only selection avoids this collapse and reaches 80.06 COMET and 21.56 BLEU, but it remains below the combined system at 81.92 COMET and 26.96 BLEU. The gap is particularly visible in BLEU, where the combined objective improves by 5.40 points over QE only. A global score can distinguish broadly adequate candidates, but a sparse pronoun error may have limited influence on the sentence-level score, especially when the remaining content is translated well. The local term changes the ordering among such candidates by assigning explicit importance to the pronoun position.

The two components therefore constrain different failure modes. QE discourages candidates that satisfy the local criterion by damaging the rest of the sentence, while pronoun feedback prevents a high global score from making the targeted decision effectively invisible. Their combination gives the best overall balance in the context-aware setting. Importantly, the combined reward does not require both component values to rise monotonically or to have directly comparable scales. Its role is to produce a useful ordering of sampled candidates, and the held-out COMET and BLEU results indicate that this ordering is more effective than either component alone.

\subsection{How Does Preceding Context Affect Targeted Feedback?}
The combined method also improves the context-agnostic model, from 56.68 to 73.95 COMET and from 7.89 to 15.26 BLEU. This result shows that the improvement is not dependent on access to a preceding-sentence cue. The complete \textsc{ProNMT} training procedure also improves the model when only the current sentence is available. At the same time, the highest overall performance is obtained when reward-guided training starts from a model adapted to the context-prefixed input. Relative to no-context \textsc{ProNMT}, the context-aware system is higher by 7.97 COMET and 11.69 BLEU.

This pattern is consistent with targeted feedback becoming more informative when the model can access evidence relevant to the targeted decision. Without preceding context, some pronoun choices remain underdetermined, and the reward can only favor agreement with the reference rather than teach a reproducible contextual mapping. With context, the sampled candidates can differ in whether they exploit the available antecedent information, giving the local reward a more meaningful basis for ranking them. The result does not establish that every filtered example requires cross-sentence resolution, nor does it directly measure context usage. It instead shows that the strongest trade-off between global quality and the targeted reward occurs in the condition where potentially relevant context is available.

The raw context baseline is not the main comparison because the base model has not yet learned the artificial context format. This issue is especially visible in English--French, where simply adding the prefix lowers COMET from 74.29 to 60.58 and BLEU from 26.23 to 16.42. After context-aware SFT, performance recovers to 76.84 COMET and 28.92 BLEU. These numbers show that access to additional text is not automatically beneficial: the model first needs to learn the structural distinction between contextual evidence and content to translate. Context-aware SFT is therefore the fair baseline for assessing the contribution of \textsc{ProNMT}, while the raw context baseline mainly quantifies the cost of introducing an unfamiliar input format.

The different baseline behavior across English--German and English--French also cautions against treating context concatenation as a universally positive architectural intervention. Its value depends on adaptation, data characteristics, and how the prefix interacts with the pretrained model. The empirical pattern is narrower and clearer: once the model can process the context-prefixed format, the subsequent \textsc{ProNMT} stage is associated with further improvements in the produced outputs.
\begin{figure*}[t]
    \centering
    \subfloat[Without context: QE reward]{\includegraphics[width=0.48\textwidth]{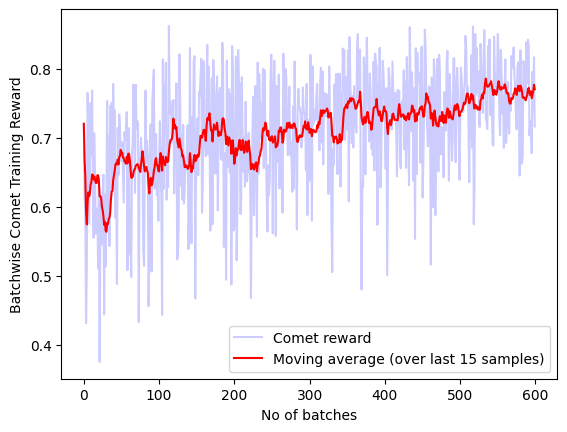}}
    \hfill
    \subfloat[Without context: pronoun reward]{\includegraphics[width=0.48\textwidth]{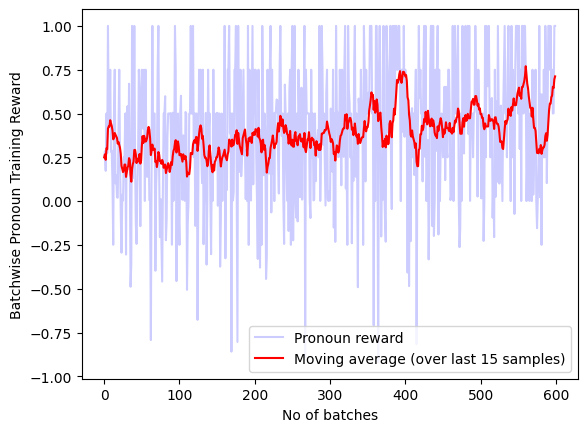}}

    \subfloat[With context: QE reward]{\includegraphics[width=0.48\textwidth]{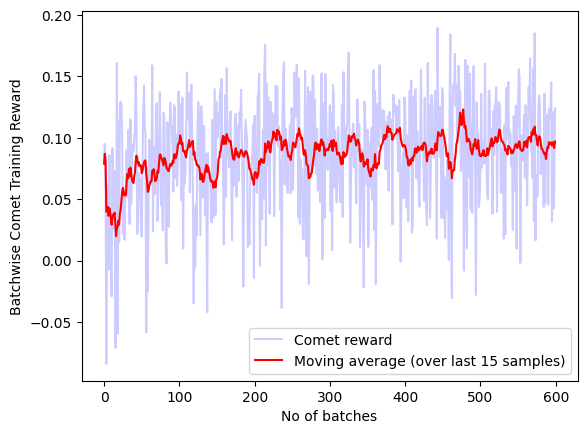}}
    \hfill
    \subfloat[With context: pronoun reward]{\includegraphics[width=0.48\textwidth]{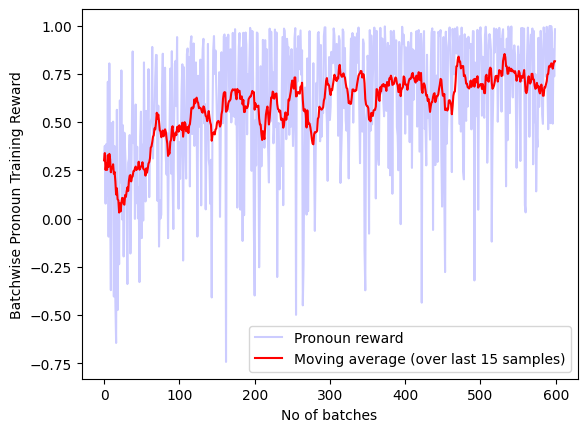}}
    \caption{Batchwise reward trajectories for combined-reward training. Curves show raw values and moving averages.}
    \label{fig:training_combined}
\end{figure*}

\subsection{Confidence Weighting versus Binary Feedback}
The binary variant obtains a larger raw pronoun reward but lower external translation scores. In context-aware English--German, binary feedback reaches 78.58 COMET and 21.58 BLEU, compared with 81.92 and 26.96 for confidence-weighted feedback. The same ordering holds without context, where the confidence-weighted system exceeds the binary variant by 10.93 COMET and 3.01 BLEU. Thus, the apparent advantage of the binary system on its own local reward does not transfer to the external metrics.

A hard $+1/-1$ signal gives every matching or non-matching candidate the same magnitude. Consequently, candidates with very different model support can become indistinguishable under the local component, and a single pronoun match can dominate small differences in QE depending on the reward scale. Pronoun-position confidence provides a finer candidate ranking because the magnitude reflects the model's confidence at the generated pronoun position. In this selection-based setting, the benefit should not be described as reducing gradient variance, since gradients are taken through the supervised update rather than through the reward. The empirical advantage is instead that confidence weighting offers more resolution when candidates are ordered before the update.

The comparison also illustrates why raw values from differently defined rewards should not be compared as though they were the same metric. A binary score is structurally encouraged to approach its extreme values, whereas confidence-weighted scores depend on the model distribution. The relevant evidence is not that one internal number is numerically larger, but that the candidate ranking induced by confidence weighting leads to better held-out translations.

\subsection{Transfer to English--French}
Table~\ref{tab:enfr_main} reports English--French results using the English--German reward weights without retuning. In the context-aware condition, \textsc{ProNMT} improves over SFT from 76.84 to 79.37 COMET and from 28.92 to 30.54 BLEU. The corresponding gains are 2.53 COMET and 1.62 BLEU. The no-context condition also improves over its baseline, from 74.29 to 76.16 COMET and from 26.23 to 28.03 BLEU.

This transfer is informative because the French setting is not a direct replication of the German pronoun inventory. English--German focuses on \,\textit{it} and the alternatives \,\textit{er}, \,\textit{sie}, and \,\textit{es}, whereas English--French focuses on \,\textit{they} and the gendered alternatives \,\textit{ils} and \,\textit{elles}. Reusing the same reward weights therefore tests whether the balance between global and local feedback remains useful when both the language pair and the targeted mapping change. The positive result suggests that the central design choice is not tied only to the numerical optimum of the English–German validation set and can transfer to a second language pair and target mapping without retuning.

The evidence is still limited to two related target languages and a single translation model, so it does not establish broad multilingual generality. It does, however, strengthen the interpretation of the English--German result. The improvement is unlikely to be solely an idiosyncrasy of one filtered corpus or one particular set of German pronouns, because the same combined objective also improves a separately constructed French setting without additional tuning. The larger COMET gain over SFT in English--French further suggests that the value of reward-guided refinement can depend on how much room remains after supervised context adaptation.

\begin{table}[t]
\centering
\caption{English--French test performance. English--German reward weights are reused without retuning. QE and ProConf are internal rewards and use a different QE model from Table~\ref{tab:ende_main}.}
\label{tab:enfr_main}
\resizebox{\columnwidth}{!}{%
\begin{tabular}{lrrrr}
\toprule
\textbf{Method} & \textbf{COMET} & \textbf{BLEU} & \textbf{QE} & \textbf{ProConf}\\
\midrule
\multicolumn{5}{l}{\textit{Without context}}\\
Baseline & 74.29 & 26.2314 & 0.7610 & 0.3197\\
\textsc{ProNMT} & \textbf{76.16} & \textbf{28.0341} & \textbf{0.7914} & \textbf{0.3846}\\
\midrule
\multicolumn{5}{l}{\textit{With context}}\\
Baseline & 60.58 & 16.4238 & 0.7571 & 0.2806\\
Context-aware SFT & 76.84 & 28.9204 & 0.7638 & 0.1678\\
\textsc{ProNMT} & \textbf{79.37} & \textbf{30.5399} & \textbf{0.8007} & \textbf{0.4908}\\
\bottomrule
\end{tabular}%
}
\end{table}

\subsection{What Does Reward-Weight Selection Reveal?}
Table~\ref{tab:weight_sensitivity} provides an additional view of the trade-off between the two reward components. In the context-aware condition, the configuration with the highest internal ProConf is $(1,1/\#\mathrm{avg\mbox{-}len})$, which reaches 0.8543, but its COMET and BLEU scores are 78.98 and 21.95. The selected configuration, $(1.2,1/\#\mathrm{tokens})$, has a substantially lower ProConf of 0.4213 but the best external scores of 80.72 COMET and 26.43 BLEU. This mirrors the component ablation: maximizing the local diagnostic is not equivalent to producing the best translation system.

Sentence-specific normalization of the pronoun term is the better choice at the selected QE weight, exceeding average-length normalization by 0.51 COMET and 1.19 BLEU at $\alpha=1.2$, though the two are within 0.32 COMET at $\alpha=1$. The comparison suggests that the balance is affected not only by the nominal values of $\alpha$ and $\beta$, but also by whether the local contribution is normalized according to each sentence. Since a pronoun-level term is combined with a sentence-level QE score, sentence-specific scaling can prevent the relative influence of the local term from changing simply because inputs differ in length.

The no-context validation results show a different optimum: $(1,1/\#\mathrm{avg\mbox{-}len})$ gives the highest COMET and BLEU, while increasing $\alpha$ to 1.2 hurts both metrics. This difference supports selecting reward weights on the validation set separately for the two input conditions rather than assuming that a single balance must be optimal everywhere. More broadly, the sensitivity table reinforces the main narrative that global and local signals are complementary but not interchangeable. Their usefulness depends on maintaining a scale at which the local term can influence candidate choice without overwhelming sentence-level quality.

\subsection{Relationship between Internal Rewards and External Metrics}
Across the ablations, internal rewards and external metrics are related but clearly not interchangeable. Pronoun-only and binary training can obtain large local values while producing lower COMET and BLEU, so the validation configuration with the highest ProConf is not the selected system. Conversely, context-aware SFT has a relatively low ProConf of 0.1406 but strong external performance. These cases show that ProConf is best understood as a description of the candidate-selection objective rather than a standalone evaluation measure.

QE is also an internal training signal, even though it is derived from a learned MT evaluation model. Optimizing the same QE model used for candidate selection may improve its score without guaranteeing equivalent gains under a separate evaluator. For that reason, the main results include reference-based COMET and BLEU, and the paper bases its conclusions on agreement between these held-out metrics rather than on the training rewards alone. The combined system's advantage is strongest when it improves both external metrics while avoiding the local-reward collapse observed in the ablations.

\subsection{Training Dynamics}
Figure~\ref{fig:training_combined} shows the batchwise QE and pronoun rewards for the English--German combined-reward runs. Both raw trajectories fluctuate substantially because each update is based on sampled candidates and a small mini-batch. The moving averages are therefore more informative than individual batch values. In the no-context run, the QE moving average generally rises over training, while the pronoun trajectory improves more gradually and remains highly variable. In the context-aware run, the pronoun moving average increases markedly during the early updates and then fluctuates around a higher range, while the QE trajectory improves early and later remains comparatively stable.

The trajectories are compatible with the two reward components learning at different rates. The targeted signal can change quickly when candidate pronouns are re-ranked, whereas sentence-level quality aggregates many decisions and may move more conservatively. At the same time, the occasional negative or low batch rewards show that sampling continues to expose difficult cases throughout training. A smooth monotonic curve is not expected because the procedure repeatedly changes the model that generates the candidate pool.

These plots are diagnostic rather than the basis for model selection. Checkpoints are selected using the combined validation reward, and final claims are drawn from held-out COMET and BLEU. This separation is important because an apparently improving training component can still correspond to poorer external quality, as demonstrated directly by the pronoun-only and binary ablations. The curves are therefore most useful for showing that optimization is active and noisy, while the tables determine whether the resulting model is actually better.

\section{Conclusion}
We presented \textsc{ProNMT}, a reward-guided iterative self-training method for balancing global translation quality with pronoun-specific feedback. The experiments support three conclusions. First, context-aware supervised fine-tuning is a strong starting point, but combined reward-guided training improves further in both English--German and English--French. Second, pronoun-only feedback is unsafe as an isolated objective because a high local reward can coincide with severe deterioration in COMET and BLEU. Third, confidence-weighted feedback provides a more useful candidate ranking than a hard binary signal. More broadly, targeted linguistic feedback is most effective when it is constrained by a global quality objective and paired with access to the information relevant to the targeted decision.

\section*{Limitations}
The study covers one model and two pronoun-focused language-pair settings. The filtered corpora identify aligned pronoun alternatives, but they do not establish that every example requires preceding context. Because the evaluation sets are constructed around the targeted pronoun phenomena, improvements in BLEU and COMET on these sets do not establish improvements on naturally distributed MT data or on sentences that do not contain the targeted phenomenon. We also do not compare against contemporary LLM-based contextual MT systems; accordingly, our experiments characterize the behavior of the proposed training objective rather than establish state-of-the-art contextual MT performance.  The method also depends on the reliability of the QE model, accurate pronoun identification, word alignment during data construction, and validation-based selection of the reward weights. Because English--German and English--French use different QE models, their raw QE values are not directly comparable. Finally, the method targets phenomena that can be represented through explicit token-position signals; it may not transfer directly to discourse properties that lack such localized supervision.

\section*{Acknowledgments}

Baban Gain acknowledges the Prime Minister's Research Fellowship (PMRF), Ministry of Education, Government of India, for supporting this research under Grant No.~2703563. The authors gratefully acknowledge support from the grant titled ``COIL-D: Centre of Indian Language Data,'' sponsored by MeitY, Government of India (Grant No. 11(14)/2018-HCC(TDIL)Voll-III-Part(21)).
\bibliography{anthology,custom}

\end{document}